\documentclass{notices}

\usepackage{amssymb,mathtools,enumerate}
\usepackage{adjustbox}

\newtheorem{definition}{Definition}

\usepackage{xcolor}
\hypersetup{
    colorlinks,
    linkcolor={red!50!black},
    citecolor={blue!50!black},
    urlcolor={blue!80!black}
}

\usepackage{tikz} 
\usetikzlibrary{cd}

\usepackage{tikz-3dplot}
\usepackage{gensymb}

\usetikzlibrary{positioning}

\usepackage{scalerel}
\usepackage{pgfplots}

\usepackage{butterfly}

\newcommand{\blue}[1]{#1}

\DeclareMathOperator{\GL}{GL}
\DeclareMathOperator{\SE}{SE}
\DeclareMathOperator{\SOr}{SO}
\DeclareMathOperator{\Or}{O}

\DeclareMathOperator{\sgn}{sgn}

\let\latexcirc=\circ
\newcommand{\ccirc}{\mathbin{\mathchoice
  {\xcirc\scriptstyle}
  {\xcirc\scriptstyle}
  {\xcirc\scriptscriptstyle}
  {\xcirc\scriptscriptstyle}
}}
\newcommand{\xcirc}[1]{\vcenter{\hbox{$#1\latexcirc$}}}
\let\circ\ccirc

\title{What is an equivariant neural network?}

\author{
Lek-Heng~Lim
  \affil{
    The first author is a professor of computational and applied mathematics at the University of Chicago. His email address is lekheng@uchicago.edu.
    }
  \and
Bradley~J.~Nelson
  \affil{
    The second author is William~H.~Kruskal Instructor of computational and applied mathematics at the University of Chicago. His email address is bradnelson@uchicago.edu.
   }
}

\begin{document}
\maketitle

We explain equivariant neural networks, a notion underlying breakthroughs in machine learning from deep convolutional neural networks for computer vision \cite{AlexNet} to AlphaFold~2 for protein structure prediction \cite{Alpha}, without assuming knowledge of equivariance or neural networks. The basic mathematical ideas are simple but are often obscured by engineering complications that come with practical realizations. We extract and focus on the mathematical aspects, and limit ourselves to a cursory treatment of the engineering issues at the end.

Let $\mathbb{V}$ and $\mathbb{W}$ be sets, and $f \colon \mathbb{V} \to \mathbb{W}$ a function.  If a group $G$ acts on both $\mathbb{V}$ and $\mathbb{W}$, and this action commutes with the function $f$:
\[
    f(x \cdot v) = x\cdot f(v) \quad \text{for all } v\in \mathbb{V},\; x \in G,
\]
then we say that $f$ is $G$-\emph{equivariant}. \blue{The special case where $G$ acts trivially on $\mathbb{W}$ is called $G$-\emph{invariant}. Linear equivariant maps are well-studied in representation theory and continuous equivariant maps are well-studied in topology. The novelty of equivariant neural networks is that they are usually neither linear nor continuous, even when  $\mathbb{V}$ and $\mathbb{W}$ are vector spaces and the actions of $G$ are linear.} 

Equivariance is ubiquitous in applications where symmetries in the input space $\mathbb{V}$ produce symmetries in the output space $\mathbb{W}$. \blue{We consider a simple example. An image may be regarded as a function $v \colon \mathbb{R}^2 \to \mathbb{R}^3$, with each pixel $p = (p_1,p_2) \in \mathbb{R}^2$ assigned some RGB color $(r,g,b) \in \mathbb{R}^3$.  A simplifying assumption here is that pixels and colors can take values in a continuum. Let $\mathbb{V} = \mathbb{W}$ be the set of all images. Let the group $G = \{1,x\} \cong \mathbb{Z}/2\mathbb{Z}$ act on $\mathbb{V}$ via \emph{top-bottom} reflection, i.e., $x\cdot v$ is the image whose value at $(p_1,p_2)$ is  $v (p_1,-p_2)$. Let $\sigma \colon \mathbb{R}^3 \to \mathbb{R}^3$,
\[
\sigma(r,g,b) = \begin{cases}
(0,0,0) & \text{if } r=g=b =0,\\
(255,255,255) & \text{otherwise}.
\end{cases}
\]
Here $(0,0,0)$ and $(255,255,255)$ are the RGB encodings for pitch black and pure white respectively. So the map $f \colon \mathbb{V} \to \mathbb{V}$, $f(v) = \sigma \circ v$ transforms a color image into a black-and-white image.}
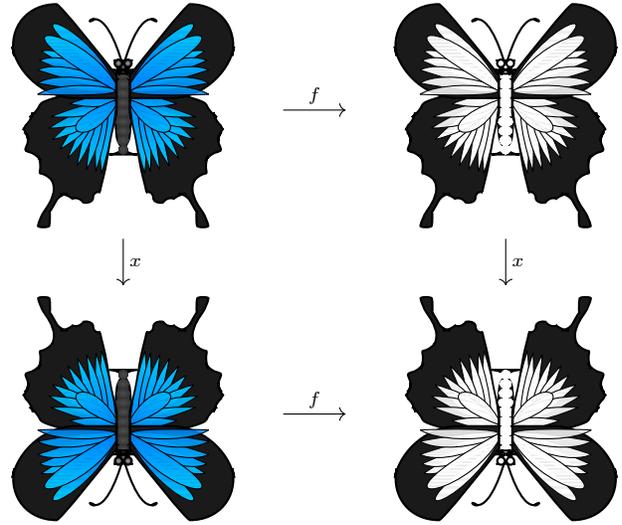
\[
\hspace*{-4ex}
\begin{tikzcd}
\begin{tikzpicture}[scale=0.25]
    \butterfly
\end{tikzpicture}
 \arrow[r, "f"] \arrow[d, "x"]
    & \begin{tikzpicture}[scale=0.25]
    \butterflyoutline
\end{tikzpicture}
 \arrow[d, "x"] \\
|[rotate=180]| \begin{tikzpicture}[scale=0.25]
    \butterfly
\end{tikzpicture}
 \arrow[r, "f"]
& |[rotate=180]| \begin{tikzpicture}[scale=0.25]
    \butterflyoutline
\end{tikzpicture}
\end{tikzcd}
\]
\blue{It does not matter whether we do a top-bottom reflection first or remove color first, the result is always the same,  i.e., $f (x \cdot v) = x \cdot f(v)$ for all $v \in \mathbb{V}$ ---  note that this holds for any color image, not just the butterfly image. Hence the decoloring map $f$ is $G$-equivariant.

Our choice of an image with left-right symmetry presents another opportunity to illustrate the notion. If we choose coordinates so that the vertical axis passes through the center of the butterfly image, then as a function $v \colon \mathbb{R}^2 \to \mathbb{R}^3$, it is invariant under the action of $H = \{1,s\}  \cong \mathbb{Z}/2\mathbb{Z}$ on $\mathbb{R}^2$ via $s(p_1,p_2) = (-p_1,p_2)$, i.e., $v(s \cdot p) = v(p)$. Note that the $G$-equivariance of $f$ has nothing to do with this.}
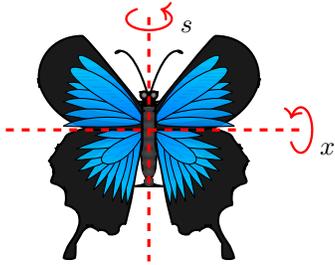
\[
\begin{tikzpicture}[scale=0.25]
    \butterfly
    \begin{scope}
    \draw [red,very thick,dashed] (0,7) -- (0,-6) node [at start] {\AxisRotator[x=1ex,y=2ex,->,rotate=-90,solid]};
    \node at (2,6.5) {$s$};
    \end{scope}
    \begin{scope}
    \draw [red,very thick,dashed] (8,1) -- (-8,1)  node [at start] {\AxisRotator[x=1ex,y=2ex,->,solid]};
    \node at (9.5,0) {$x$};
    \end{scope}
\end{tikzpicture}
\]

\blue{A takeaway of these examples is that nonlinear and discontinous functions may very well be equivariant. However, the best known context for discussing equivariant maps is when $f$ is an \emph{intertwining operator}, i.e., a linear map between vectors spaces} $\mathbb{V}$ and $\mathbb{W}$ equipped with a linear action of $G$. In this case, an equivalent formulation of $G$-equivariance takes the following form: Given linear representations of $G$ on $\mathbb{V}$ and $\mathbb{W}$, i.e., homomorphisms $\rho_1 \colon G \to \GL(\mathbb{V})$ and $\rho_2 \colon G\to \GL(\mathbb{W})$,  a linear map $f \colon \mathbb{V} \to \mathbb{W}$ is said to be $G$-equivariant if
\begin{equation}\label{eq:linear}
f\bigl(\rho_1(x) v\bigr) = \rho_2(x) f(v)  \quad \text{for all } v\in \mathbb{V},\; x \in G.
\end{equation}
\blue{Intertwining operators preserve eigenvalues and, when $G$ is a Lie group, the action of its Lie algebra, properties that are crucial to their use in physics \cite{Baez}.}

Nevertheless the restriction to linear maps is unnecessary.  The \emph{de~Rham problem} asks if $\mathbb{V} = \mathbb{W} = \mathbb{R}^n$ and $f \colon \mathbb{R}^n \to \mathbb{R}^n$ is merely required to be a homeomorphism, then does the condition \eqref{eq:linear} implies that $f$ must be a linear map? De~Rham conjectured this to be the case but it was disproved in \cite{Cappell}, launching a fruitful study of \emph{nonlinear similarity}, i.e., nonlinear homeomorphisms $f$ with
\[
f \rho_1(x) f^{-1} = \rho_2(x) \quad \text{for all }  x \in G,
\]
in algebraic topology and algebraic K-theory, \blue{and more generally the study of equivariant continuous maps in equivariant topology \cite{Gerhardt,Tu}.}

An equivariant neural network \cite{Welling} \blue{is an equivariant map $f$} constructed from alternately composing equivariant linear maps with nonlinear ones \blue{like the decoloring map above}. That neural networks can be readily made equivariant is a consequence of two \blue{straightforward} observations:
\begin{enumerate}[\upshape (i)]
    \item the composition of two $G$-equivariant functions $f \colon \mathbb{V}\to \mathbb{W}$,  $g \colon  \mathbb{U}\to \mathbb{V}$ is $G$-equivariant;
    \item the linear combination of two $G$-equivariant functions $f,g \colon \mathbb{V}\to \mathbb{W}$ is $G$-equivariant;
\end{enumerate}
\blue{even when $f,g$ are nonlinear. Although an equivariant neural network is nonlinear, it uses intertwining operators as building blocks, and \eqref{eq:linear} plays a key role.} 
In some applications like \blue{the wave function on p.~\pageref{pg:wave} or in} \cite{zaheer_deep_2017}, the input $\mathbb{V}$ or possibly some hidden layers may not be vector spaces; for simplicity we assume that they are and their $G$-actions are linear. 

In machine learning applications, the \blue{map} $f$ is  learned from data. A major advantage of requiring equivariance in a neural network $f$ is that it allows one to greatly narrow down the search space for the parameters that define $f$.  To demonstrate this, we begin with a simplified case that avoids group representations. A feed-forward neural network is a function $f \colon \mathbb{R}^n \to \mathbb{R}^n$
 obtained by alternately composing affine maps $\alpha_i \colon \mathbb{R}^n \to \mathbb{R}^n $, $i=1,\dots,k$, with a nonlinear function $\sigma \colon  \mathbb{R}^n \to \mathbb{R}^n$, i.e.,
\[
\mathbb{R}^n \xrightarrow{\alpha_1} \mathbb{R}^n \xrightarrow{\sigma} \mathbb{R}^n \xrightarrow{\alpha_2} \mathbb{R}^n \xrightarrow{\sigma}
 \cdots  \xrightarrow{\sigma} \mathbb{R}^n \xrightarrow{\alpha_k} \mathbb{R}^n,
\]
giving $f = \alpha_k \circ \sigma \circ \alpha_{k -1} \circ \cdots \circ \sigma \circ \alpha_2 \circ \sigma \circ \alpha_1$.
The \emph{depth}, also known as the number of \emph{layers}, is $k$ and the \emph{width}, also known as the number of \emph{neurons}, is $n$. The simplifying assumption, which will be dropped later, is that our neural network has constant width throughout all layers. The nonlinear function $\sigma$ is called an \emph{activation}, with the ReLU (rectified linear unit) function
$\sigma(t) \coloneqq \max(t,0)$ for $t \in \mathbb{R}$ a standard choice. In a slight abuse of notation, the activation is extended to vector inputs \blue{$v = (v_1,\dots,v_n) \in \mathbb{R}^n$ by evaluating} coordinatewise
\begin{equation}\label{eq:ptwise}
\sigma(v) = (\sigma(v_1),\dots,\sigma(v_n))
\end{equation}
In this sense, $\sigma \colon  \mathbb{R}^n \to \mathbb{R}^n$ is called a \emph{pointwise nonlinearity}.
The affine function is defined by $\alpha_i(v) = A_iv + b_i$ for some $A_i \in \mathbb{R}^{n \times n}$ called the \emph{weight} matrix and some $b_i \in \mathbb{R}^n$ called the \emph{bias} vector. We do not include a bias $b_k$ in the last layer.

Although convenient, it is somewhat misguided to lump the bias and weight together in an affine function. Each bias $b_i$ is intended to serve as a \emph{threshold} for the activation $\sigma$ and should be part of it, detached from the weight $A_i$ that transforms the input. If one would like to incorporate translations, one may do so by going up one dimension, observing that $\begin{bsmallmatrix} A & b \\ 0 &  1 \end{bsmallmatrix}\begin{bsmallmatrix} v  \\ 1 \end{bsmallmatrix} = \begin{bsmallmatrix} Av + b  \\ 1 \end{bsmallmatrix}$.
Hence, a better, but mathematically equivalent, description of $f$ would be as the composition
\begin{multline*}
\mathbb{R}^n \xrightarrow{A_1}  \mathbb{R}^n \xrightarrow{\sigma_{b_1}}  \mathbb{R}^n \xrightarrow{A_2}  \mathbb{R}^n \xrightarrow{\sigma_{b_2}}  \\
\cdots  \xrightarrow{\sigma_{b_{k-1}}}  \mathbb{R}^n \xrightarrow{A_k}  \mathbb{R}^n
\end{multline*}
where we identify $A_i \in \mathbb{R}^{n \times n}$ with the linear operator $\mathbb{R}^n \to \mathbb{R}^n$, $v \mapsto A_i v$, and for any $b \in \mathbb{R}^n$ we define $\sigma_b \colon \mathbb{R}^n \to \mathbb{R}^n$ by $\sigma_b(v) = \sigma(v+b)\in \mathbb{R}^n$. We will drop the composition symbol $\circ$ to avoid clutter and write
\[
f = A_k \sigma_{b_{k-1}}  A_{k -1}  \cdots  \sigma_{b_2} A_2  \sigma_{b_1}  A_1
\]
as if it is a product of matrices. For a fixed  $\theta \in \mathbb{R}$,
\begin{equation}\label{eq:thres}
\sigma_\theta(t) = \begin{cases} t -\theta & t \ge \theta, \\ 0 & t < \theta, \end{cases}
\end{equation}
plays the role of a threshold for activation as was intended in \cite[p.~392]{Rosenblatt} and \cite[p.~120]{MP}.

A major computational issue with neural networks is the large number of unknown parameters, namely the $kn^2 + (k-1)n$ entries of the weights and biases, that have to be fit with data, especially for wide neural networks where $n$ is  large.  To get an idea of the numbers involved in realistic situations, $n$ may be on the order of millions of pixels for image-based tasks, whereas $k$ is typically $10$ to $50$ layers deep. Computational cost aside, one may not even have enough data to fit so many parameters. Thus, many successful applications of neural networks require that we identify, based on the problem at hand, an appropriate low-dimensional subset of $\mathbb{R}^{n \times n}$ from which we will find our weights $A_1,\dots,A_k$. For example, for a signal processing problem, we might restrict $A_1,\dots,A_k$ to be Toeplitz matrices; the convolutional neural networks for image recognition in  \cite{AlexNet}, an article that launched the deep learning revolution, essentially restrict $A_1,\dots,A_k$ to so called block-Toeplitz--Toeplitz-block or BTTB matrices. \blue{A $k$-layer  width-$n$ neural network all of whose weight matrices are Toeplitz would require just $k(2n-1) + (k-1)n$ parameters; one weighted with BTTB matrices, i.e., $m_1 \times m_1$ block Toeplitz matrices whose blocks are $m_2 \times m_2$ Toeplitz matrices, would require $k(2m_1 - 1)(2m_2 -1) + (k-1)m_1m_2$ parameters. In either case we avoided the $n^2$ exponent, bearing in mind that in the BTTB case $n = m_1m_2$.} It turns out that convolutional neural networks are a quintessential example of equivariant neural networks \cite{Welling}, and in fact every equivariant neural network may be regarded as a generalized convolutional neural network in an appropriate sense \cite{Risi1}.

To see how equivariance naturally restricts the range of possible $A_1,\dots,A_k$, let $G \subseteq \mathbb{R}^{n \times n}$ be a matrix group.  Then $f \colon \mathbb{R}^n \to \mathbb{R}^n$ is $G$-equivariant if
\begin{equation}\label{eq:ENN0}
f(Xv) = X f(v)  \quad \text{for all } v\in \mathbb{R}^n,\; X \in G;
\end{equation}
and an equivariant neural network is simply a feed-forward neural network $f \colon \mathbb{R}^n \to \mathbb{R}^n$ that satisfies \eqref{eq:ENN0}. The key to \blue{its construction is just that}
\begin{align*}
f(Xv) &= A_k \sigma_{b_{k-1}}  A_{k -1}  \cdots  \sigma_{b_2} A_2  \sigma_{b_1}  A_1 Xv\\*
&=X (X^{-1}A_k X) (X^{-1}\sigma_{b_{k-1}} X) (X^{-1} A_{k -1} X)  \\*
&\qquad \cdots  (X^{-1} A_2 X)(X^{-1} \sigma_{b_1} X) (X^{-1} A_1 X) v\\*
&= XA_k' \sigma_{b_{k-1}}'  A_{k -1}'  \cdots  \sigma_{b_2}' A_2'  \sigma_{b_1}'  A_1' v
\end{align*}
and the last expression equals $X f(v)$ if we have
\begin{equation}\label{eq:ENN1}
A_i'  = X^{-1} A_i X = A_i, \quad \sigma_{b_i}' = X^{-1} \sigma_{b_i} X = \sigma_{b_i}
\end{equation}
for all $i=1,\dots,k$, and for all $X \in G$. The condition on the right is satisfied by \blue{any pointwise nonlinearity that takes the form in \eqref{eq:thres}, i.e., $b_i \in \mathbb{R}^n$ has all coordinates equal to some $\theta \in \mathbb{R}$; we will elaborate on this later}. The condition on the left limits the possible weights for $f$ to a (generally) much smaller subspace of matrices  that commute with all elements of $G$. Finding this subspace (in fact a subalgebra) of \blue{intertwining operators},
\begin{equation}\label{eq:intertwining}
\{A \in \mathbb{R}^{n \times n} \colon AX =XA \text{ for all } X \in G \},
\end{equation}
is a well-studied problem in group representation theory; a general purpose approach is to compute the null space of a matrix built from the generators of $G$ and, if continuous, its Lie algebra \cite{finzi_practical_2021}. We caution the reader that $G$ will generally be a very low-dimensional subset of  $\mathbb{R}^{n \times n}$, as will become obvious from our example below in \eqref{eq:com}. It will be pointless to pick, say, $G = \SOr(n)$ as the set in \eqref{eq:intertwining} will then be just $\{\lambda I \in \mathbb{R}^{n \times n}  \colon \lambda \in \mathbb{R} \}$, clearly too small to serve as meaningful weights for any neural network. Indeed, $G$ will usually be a homomorphic image of a representation $\rho \colon G \to \GL(n)$, i.e., the image $\rho(G)$ will play the role of $G$ in \eqref{eq:intertwining}. In any case, we will need to bring in group representations to address a different issue.

In general, neural networks  have \blue{different width} $n_i$ in each layer:
\begin{multline*}
\mathbb{R}^{n_0} \xrightarrow{A_1} \mathbb{R}^{n_1} \xrightarrow{\sigma_{b_1}} \mathbb{R}^{n_1} \xrightarrow{A_2} \mathbb{R}^{n_2} \xrightarrow{\sigma_{b_2}}  \cdots\\
\cdots \xrightarrow{\sigma_{b_{k-1}}} \mathbb{R}^{n_{k-1}} \xrightarrow{A_k} \mathbb{R}^{n_k}
\end{multline*}
with $A_i \in \mathbb{R}^{n_{i-1} \times n_i}$, $i =1,\dots,k$, $b_i \in \mathbb{R}^{n_i}$, $i =1,\dots,k-1$. The simplified case treated above assumes that $n_0 =n_1 = \dots =n_k = n$. It is easy to accommodate this slight complication by introducing group representations to equip every layer with its own homomorphic copy of $G$. Instead of fixing $G$ to be some subgroup of $\GL(n)$, $G$ may now be any abstract group but we introduce a homomorphism
\[
\rho_i \colon G \to \GL(n_i), \quad i =0,1,\dots,k,
\]
in each layer, and replace the equivariant condition \eqref{eq:ENN1} with the more general \eqref{eq:linear}, i.e.,
\[
\rho_i(x)^{-1} A_i \rho_{i-1}(x) = A_i, \quad \rho_i(x)^{-1} \sigma_{b_i} \rho_{i}(x) =  \sigma_{b_i}
\]
or, equivalently,
\begin{equation}\label{eq:ENN2}
A_i \rho_{i-1}(x) = \rho_i(x)A_i, \quad \sigma_{b_i} \rho_{i}(x) = \rho_i(x) \sigma_{b_i}
\end{equation}
for all $x \in G$. In case \eqref{eq:ENN2} evokes memories of Schur's Lemma, we would like to stress that the representations $\rho_i$ are in general very far from being irreducible and that the map $\sigma_{b_i}$ is nonlinear. Indeed the scenario described by Schur's Lemma is undesirable for equivariant neural networks: As we pointed out earlier, we do not want to restrict our weight matrices to the form $A_i = \lambda I$ or a direct sum of these.

We summarize our discussion with a formal definition.
\begin{definition}
Let $A_i \in \mathbb{R}^{n_{i-1} \times n_i}$, $i =1,\dots, k$, $b_i \in \mathbb{R}^{n_i}$, $i =1,\dots, k-1$, and $\sigma \colon \mathbb{R} \to \mathbb{R}$ be a continuous function. Let $G$ be a group and $\rho_i \colon G \to \GL(n_i)$, $i=0,\dots,k$, be its representations. The $k$-layer feed-forward neural network $f \colon \mathbb{R}^{n_0} \to \mathbb{R}^{n_k}$ given by
\[
f(v) = A_k \sigma_{b_{k-1}}  A_{k -1}  \cdots  \sigma_{b_2} A_2  \sigma_{b_1}  A_1 v
\]
is a $G$-\emph{equivariant neural network} with respect to $\rho_0,\dots,\rho_k$ if \eqref{eq:ENN2} holds for all $x \in G$. Here $\sigma_b \colon \mathbb{R}^{n_i} \to \mathbb{R}^{n_i}$, $\sigma_b(v) = \sigma(v+b)$, is a pointwise nonlinearity as in \eqref{eq:ptwise}.
\end{definition}
\begin{figure*}[h]
\centering
\input{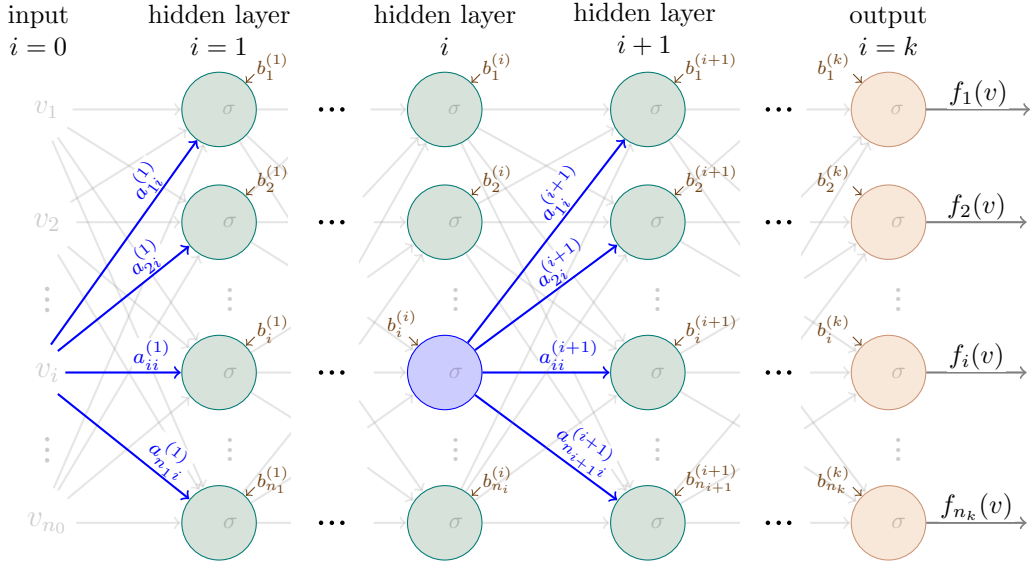}
\vspace*{-4ex}
\caption{A $k$-layer $\sigma$-activated feed-forward neural network, also known as a multilayer perceptron.}
\label{fig:nn}
\end{figure*}
\blue{A word of caution is in order here. What we call a neural network, i.e., the alternate composition of activations with affine maps, is sometimes also called a \emph{multilayer perceptron}; a standard depiction is shown in Figure~\ref{fig:nn}. When it is fitted with data, one would invariably feed its output into a \emph{loss function} and that is usually not equivariant; or one might chain together multiple units of multilayer perceptrons into larger frameworks like autoencoders, generative adversarial networks, transformers, etc, that contain other nonequivariant components. In the literature, the term ``neural network'' sometimes refers to the entire framework collectively. In our article, it just refers to the multilayer perceptron  --- this is the part that is equivariant.}

We will use \blue{an insightful toy} example as illustration. Let $\mathbb{V} = (\mathbb{R}^3)^m = \mathbb{R}^{3m}$ be the set of possible positions of $m$ unit-weight masses, $\mathbb{W} = \mathbb{R}^3$, and $f: \mathbb{V}\to \mathbb{W}$ compute the  center of mass
\begin{equation}\label{eq:com}
    f( y_1,\dots,y_m) = \frac{1}{m} \sum_{i=1}^m y_i
\end{equation}
with $y_1,\dots, y_m \in \mathbb{R}^3$. We use the same system of coordinates for each copy of $\mathbb{R}^3$ in $\mathbb{V}$ and $\mathbb{W}$.  If we work in a different coordinate system, the position of the center of mass remains unchanged but its coordinates will change accordingly.  For simplicity, we consider a linear change of coordinates, represented by the action of a matrix $X\in \GL(3)$ on each point in $\mathbb{R}^3$. By linearity,
\[
f\bigl(X(y_1,\dots,y_m)\bigr) = \frac{1}{m} \sum_{i=1}^m X y_i = X f(y_1,\dots,y_m),
\]
so $f$ is $\GL(3)$-equivariant.  Since each mass has the same unit weight, $f$ is also invariant under permutations of the input points.  Let $\pi \in S_m$, which acts on  $\mathbb{V}$ via $\pi  (y_1,\dots,y_m) = (y_{\pi(1)},\dots, y_{\pi(m)})$ and acts trivially on $\mathbb{W}$ via $\pi (y) =  y$.  As the sum in \eqref{eq:com} is permutation invariant,
\[
f\bigl(\pi(y_1,\dots,y_m)\bigr) = \pi f(y_1,\dots,y_m),
\]
so $f$ is $S_m$-invariant. Combining our two group actions, we see that $f$ is $(\GL(3) \times S_m)$-equivariant. Note that the group here is $G = \GL(3) \times S_m$, which has much lower dimension than $\GL(\mathbb{V}) = \GL(3m)$ for large $m$. This is typical in equivariant neural networks.

In this simple example, we not only know $f$ but have an explicit expression for it.  In general, there are many functions that we know should be equivariant or invariant to certain group actions, but for which we do not know any simple closed-form expression; and this is where it helps to assume that $f$ is given by some neural network whose parameters could be determined by fitting it with  data\blue{, or, if it is used as an ansatz, by plugging into some differential or integral equations.} A simple data fitting example is provided by semantic segmentation in images, which seeks to classify pixels as belonging to one of several types of objects. If we rotate or mirror an image, we expect that pixel labels should follow the pixels.
\blue{A more realistic version of the center of mass example would be} a molecule represented by positions of its atoms, which comes up in chemical property or drug response predictions. Here we want equivariance with respect to coordinate transformations, but we wish to preserve pairwise distances between atoms and chirality, so the natural group to use is $\SOr(3)$  \cite{Risi2} or the special Euclidean group $\SE(3)$ \cite{weiler_3d_2018,Fuchs}.  The much-publicized protein structure prediction engine of DeepMind’s AlphaFold~2 relies on an $\SE(3)$-equivariant neural network and an $\SE(3)$-invariant attention module \cite{Alpha}. In \cite{townshend_RNA_2021}, $\SE(3)$-equivariant convolution is used to improve accuracy assessments of RNA structure models.

\blue{Another straightforward example comes from computational quantum chemistry, where one seeks a solution to a Schr\"odinger equation: if we write $v_i \in \mathbb{R}^3 \times \{-\frac12, \frac12\}$, then the wave function of $m$ identical spin-$\frac12$ fermions is antisymmetric, i.e.,\label{pg:wave}
\[
f(v_{\pi(1)}, v_{\pi(2)}, \dots, v_{\pi(m)}) = (-1)^{\sgn(\pi)} f(v_1, v_2, \dots, v_m).
\]
for all $\pi \in S_m$. In other words, the increasingly popular  antisymmetric neural networks  \cite{PauliNet} are $S_m$-equivariant neural networks. Even without going into the details, the reader could well imagine that restricting to neural networks that are antisymmetric is a savings from having to consider all possible neural networks.} More esoteric examples in particle physics call for Lorentz groups of various stripes like $\Or(1,3)$, $\Or^+(1,3)$, $\SOr(1,3)$, or $\SOr^+(1,3)$, which are used in Lorentz-equivariant neural networks to identify top quarks in data from high-energy physics experiments \cite{Lorentz}.

\blue{We now discuss the equivariant condition  for pointwise nonlinearities $ X^{-1} \sigma_b  X = \sigma_b$. It is instructive to look at a simple numerical example. Suppose we apply a pointwise nonlinearity $\sigma$ and a permutation matrix $X$ given by
\[
\sigma(t) = \begin{cases}
+1 & t \ge 3.0,\\
-1 & t < 3.0,
\end{cases} \qquad
X = \begin{bmatrix} 0 & 1 & 0 \\ 0 & 0 & 1\\ 1 & 0 & 0 \end{bmatrix},
\]
to a vector $v =(2.1,3.4,0.2) \in \mathbb{R}^3$. We see that $ \sigma( Xv)= X\sigma (v)$:
\[
\adjustbox{max width=\linewidth}{$
\begin{matrix}
v_1 \\
v_2 \\
v_3
\end{matrix}
\begin{bmatrix}
2.1 \\
3.4 \\
0.2
\end{bmatrix} \xrightarrow{X} 
\begin{matrix}
v_{\pi(1)}  \\
v_{\pi(2)}  \\
v_{\pi(3)} 
\end{matrix}
\begin{bmatrix}
3.4 \\
0.2 \\
2.1
\end{bmatrix} \xrightarrow{\;\sigma\;}
\begin{matrix}
\sigma(v_{\pi(1)})  \\
\sigma(v_{\pi(2)})  \\
\sigma(v_{\pi(3)})
\end{matrix}
\begin{bmatrix}
+1 \\
-1 \\
-1
\end{bmatrix} \xrightarrow{X^{-1}}
\begin{matrix}
\sigma(v_1) \\
\sigma(v_2) \\
\sigma(v_3)
\end{matrix}
\begin{bmatrix}
-1 \\
+1 \\
-1
\end{bmatrix}$}
\]
which clearly holds more generally,  i.e., $X^{-1}  \sigma X= \sigma$ for any permutation matrix $X$ and any pointwise nonlinearity $\sigma$. The bottom line is that the permutation matrix $X = \rho(\pi)$ comes from a representation $\rho : S_n \to \GL(n)$; and since $\pi \in S_n$ acts on the indices of $v$ and $\sigma \colon \mathbb{R}^n \to \mathbb{R}^n$ acts on the values of $v$, the two actions are always independent of each other. More generally, it is easy to see that if we include a bias term $b \in \mathbb{R}^n$, then $\sigma_b : \mathbb{R}^n \to \mathbb{R}^n$ is $S_n$-equivariant as long as $b$ has all coordinates equal \cite{Welling}, i.e., it has the form in \eqref{eq:thres}. This does not necessarily hold for more general $b$: Take the example above and set the bias to be $b = (-1,0,0)$.
\[
\adjustbox{max width=\linewidth}{$
\begin{matrix}
v_1 \\
v_2 \\
v_3
\end{matrix}
\begin{bmatrix}
2.1 \\
3.4 \\
0.2
\end{bmatrix} \xrightarrow{X} 
\begin{matrix}
v_{\pi(1)}  \\
v_{\pi(2)}  \\
v_{\pi(3)} 
\end{matrix}
\begin{bmatrix}
3.4 \\
0.2 \\
2.1
\end{bmatrix} \xrightarrow{\;\sigma_b\;}
\begin{matrix}
\sigma_b(v_{\pi(1)})  \\
\sigma_b(v_{\pi(2)})  \\
\sigma_b(v_{\pi(3)})
\end{matrix}
\begin{bmatrix}
-1 \\
-1 \\
-1
\end{bmatrix} \xrightarrow{X^{-1}}
\begin{matrix}
\sigma_b(v_1) \\
\sigma_b(v_2) \\
\sigma_b(v_3)
\end{matrix}
\begin{bmatrix}
-1 \\
-1 \\
-1
\end{bmatrix}$}
\]
but $\sigma_b(v) = \sigma(v + b) = (-1,+1,-1)$. So $X^{-1}  \sigma_b X \ne \sigma_b$. Going beyond pointwise nonlinearity is a nontrivial issue and is crucial when the neural network requires more than just $S_n$-equivariance. We will say a few words about this below.

The mathematical ideas that we have described are all fairly straightforward. Indeed the technical challenges in equivariant neural networks are mostly about getting these mathematical ideas to work in real-life situtations, what we have swept under ``engineering complications'' rug. We will discuss a few of these but as engineering complications go, they invariably depend on the problem at hand and every case is different.

The butterfly image example presented at the beginning already concelaed several difficulties. While we have assumed that images are functions $v \colon \mathbb{R}^2 \to \mathbb{R}^3$, in real life they are sampled on a grid, i.e., pixels are discrete, and a more realistic model would be $v \colon \mathbb{Z}^2 \to \mathbb{R}^3$. Instead of a straightforward $\SOr(2)$-equivariance as one might expect for imaging problems, one instead finds discussions of equivariance \cite{Welling} with respect to unusual groups like
\[
G_1 = \biggl\{ \begin{bsmallmatrix} 1 & 0 & m_1 \\ 0 &  1 & m_2 \\  0 & 0 & 1\end{bsmallmatrix} \in \mathbb{R}^{3 \times 3} \colon  m_1,m_2 \in \mathbb{Z} \biggr\}
\]
for translation in $\mathbb{Z}^2$; or
\begin{multline*}
G_2 = \biggl\{ \begin{bsmallmatrix} \cos(k\pi/2) & -\sin(k \pi/2) & m_1 \\  \sin(k\pi/2)  &   \cos(k\pi/2)  & m_2 \\  0 & 0 & 1\end{bsmallmatrix} \in \mathbb{R}^{3 \times 3} \colon
\\ k =0,1,2,3;\;  m_1,m_2 \in \mathbb{Z} \biggr\}
\end{multline*}
that augments $G_1$ with right-angle rotations; or
\begin{multline*}
G_3 = \biggl\{ \begin{bsmallmatrix} (-1)^j \cos(k\pi/2) & (-1)^{j+1} \sin(k \pi/2) & m_1 \\  \sin(k\pi/2)  &   \cos(k\pi/2)  & m_2 \\  0 & 0 & 1\end{bsmallmatrix} \in \mathbb{R}^{3 \times 3} \colon \\ \begin{multlined} k =0,1,2,3; \\  j=0,1; \;  m_1,m_2 \in \mathbb{Z} \end{multlined}\biggr\}
\end{multline*}
that further augments $G_2$ with reflections. The reason for these choices is that they have to send pixels to pixels.

There is also the important issue of \emph{aliasing} \cite{Zhang}. When pixels are discrete, rotation will involve interpolation, and pointwise nonlinearities introduce higher order harmonics that produce aliasing and break equivariance \cite{Franzen}. This 
can happen even with discrete translations like those in the groups $G_1,G_2,G_3$ above for standard convolutional neural networks. Dealing with aliasing  and choosing equivariant activations that do not compromise expressive power \cite{Maron} are important problems that cannot be underemphasized; and dealing with these issues constitute a mainstay of the research and development in equivariance neural networks.

In fact, in reality the pixels of an image are not just discrete but also finite in number. So instead of  $v \colon \mathbb{Z}^2 \to \mathbb{R}^3$ a $p$-pixel image is more accurately a function $v \colon \{x_1,\dots,x_p\} \to \mathbb{R}^3$ on some discrete finite subset of points $x_1,\dots,x_p\in \mathbb{R}^2$. Since these $p$ points are fixed, we may conveniently regard $v \in \mathbb{R}^p \oplus \mathbb{R}^p \oplus \mathbb{R}^p$ with each copy of $\mathbb{R}^p$ representing one of  three color \emph{channels}. In such cases the output of each layer should not be simply treated} as a vector space $\mathbb{R}^{n_i}$ but a direct sum $\mathbb{R}^{n_i} = \mathbb{R}^{p_1} \oplus \dots \oplus \mathbb{R}^{p_m}$, with $p_1,\dots,p_m$ depending on $i$ and $n_i = p_1 + \dots + p_m$. The matrix $A_i : \mathbb{R}^{n_{i-1}} \to \mathbb{R}^{n_i}$ would then have a corresponding block structure and the representation $\rho_i: G \to \GL(n_i)$ takes the form  $\rho_i = \bigoplus_{j=1}^m \rho_{ij}$ with  $\rho_{ij} : G \to \GL(p_j)$.

\blue{For molecular structure prediction problems, in} \cite{Fuchs,townshend_RNA_2021}, the input is a collection of points $y_1,\dots,y_m$ augmented with various information in addition to location coordinates, giving the input layer $\mathbb{R}^{n_0}$ a direct sum structure $\mathbb{R}^{p_1} \oplus \dots \oplus \mathbb{R}^{p_m}$ that propagates through later layers. Just to give a flavor of what is involved, in \cite{townshend_RNA_2021}, the inputs $y_1,\dots,y_m$ are atom positions in a model of an RNA molecule, with an encoding of atom type, and the output is an estimate of the root mean square error of the model's structure; in one example in \cite{Fuchs}, the inputs $y_1,\dots,y_m$ encode position, velocity, and charge of $m$ particles, and the output is an estimate of the location and velocity of each particle after some amount of time. In both examples,  the  function $f$ that maps inputs to outputs has no known expression but is known to be equivariant with respect to $\SE(3)$, i.e., translations and rotations of the coordinate system. The matrix $A :  \mathbb{R}^{q_1} \oplus \dots \oplus \mathbb{R}^{q_m} \to  \mathbb{R}^{p_1} \oplus \dots \oplus \mathbb{R}^{p_m}$ has a block structure $A = [A_{ij}]_{i,j=1}^m$, $A_{ij} \in \mathbb{R}^{p_i \times q_j}$, and is equivariant if each block $A_{ij} : \mathbb{R}^{q_j} \to \mathbb{R}^{p_i}$ is equivariant.
Equivariance constrains each block $A_{ij}$ to depend entirely on the relative input locations $y_i - y_j$, and the permitted matrices can be expressed in terms of radial kernels, spherical harmonics, and Clebsch--Gordan coefficients \cite{weiler_3d_2018}.

The engineering aspects of equivariant neural networks are many and varied. While we have selectively discussed a few that are more common and mathematical in nature, we have also ignored many that are specific to the application at hand and often messy. We avoided most jargon used in the original literature as it tends to be mathematically imprecise or application specific. Nevertheless, we stress that equivariant neural networks are ultimately used in an engineering context and a large part of their success has to do with overcoming real engineering challenges.

\subsection*{Acknowledgment}
\blue{We would like to thank the two anonymous reviewers for their many useful suggestions and comments.} We would also like to thank  Risi Kondor, \blue{Zehua Lai,} and Jiangying Zhou for helpful discussions. Both authors are supported by DARPA HR00112190040; LHL is supported by NSF DMS 1854831 \blue{and ECCS 2216912}; and BN is supported by NSF DMS 1547396.
\blue{The butterfly figure used on the first page was originally created by Ulysses.} The simplified constant width case was adapted from \cite[Example~2.16]{L} where it was used to illustrate tensor transformation rules.

\bibliographystyle{natabbrv}
\bibliography{equi.bib}

\begin{bibdiv}
\begin{biblist}

\bib{Lorentz}{inproceedings}{
      author={Bogatskiy, Alexander},
      author={Anderson, Brandon},
      author={Offermann, Jan},
      author={Roussi, Marwah},
      author={Miller, David},
      author={Kondor, Risi},
       title={{L}orentz group equivariant neural network for particle physics},
        date={2020},
   booktitle={{Proceedings of the 37th International Conference on Machine
  Learning (ICML)}},
      editor={{Daum\'e III}, Hal},
      editor={Singh, Aarti},
      series={Proceedings of Machine Learning Research},
      volume={119},
       pages={992\ndash 1002},
}

\bib{Baez}{article}{
      author={Baez, John},
      author={Huerta, John},
       title={The algebra of grand unified theories},
        date={2010},
     journal={Bull. Amer. Math. Soc. (N.S.)},
      volume={47},
      number={3},
       pages={483\ndash 552},
         url={https://doi.org/10.1090/S0273-0979-10-01294-2},
}

\bib{Cappell}{article}{
      author={Cappell, Sylvain~E.},
      author={Shaneson, Julius~L.},
       title={Nonlinear similarity},
        date={1981},
     journal={Ann. of Math.},
      volume={113},
      number={2},
       pages={315\ndash 355},
         url={https://doi.org/10.2307/2006986},
}

\bib{Welling}{inproceedings}{
      author={Cohen, Taco},
      author={Welling, Max},
       title={Group equivariant convolutional networks},
        date={2016},
   booktitle={{Proceedings of the 33rd International Conference on Machine
  Learning (ICML)}},
      editor={Balcan, Maria~Florina},
      editor={Weinberger, Kilian~Q.},
      series={Proceedings of Machine Learning Research},
      volume={48},
       pages={2990\ndash 2999},
}

\bib{Franzen}{inproceedings}{
      author={Franzen, Daniel},
      author={Wand, Michael},
       title={General nonlinearities in {$\SOr(2)$}-equivariant cnns},
        date={2021},
   booktitle={{Advances in Neural Information Processing Systems (NeurIPS)}},
      editor={Ranzato, Marc'Aurelio},
      editor={others},
      volume={34},
       pages={1970\ndash 1981},
}

\bib{Fuchs}{inproceedings}{
      author={Fuchs, Fabian},
      author={Worrall, Daniel~E.},
      author={Fischer, Volker},
      author={Welling, Max},
       title={{$\SE(3)$}-transformers: {3D} roto-translation equivariant
  attention networks},
        date={2020},
   booktitle={{Advances in Neural Information Processing Systems (NeurIPS)}},
      editor={Larochelle, Hugo},
      editor={others},
       pages={1970\ndash 1981},
}

\bib{finzi_practical_2021}{inproceedings}{
      author={Finzi, Marc},
      author={Welling, Max},
      author={Wilson, Andrew~Gordon},
       title={A practical method for constructing equivariant multilayer
  perceptrons for arbitrary matrix groups},
        date={2021},
   booktitle={{Proceedings of the 38th International Conference on Machine
  Learning (ICML)}},
       pages={3318\ndash 3328},
}

\bib{Gerhardt}{article}{
      author={Gerhardt, Teena},
       title={Invariants of rings via equivariant homotopy},
        date={2019},
     journal={Notices Amer. Math. Soc.},
      volume={66},
      number={8},
       pages={1353\ndash 1354},
}

\bib{PauliNet}{article}{
      author={Hermann, Jan},
      author={Sch\"atzle, Zeno},
      author={No\'e, Frank},
       title={Deep-neural-network solution of the electronic schr\"odinger
  equation},
        date={2020},
     journal={Nat. Chem.},
      volume={12},
       pages={891\ndash 897},
         url={https://doi.org/10.1038/s41557-020-0544-y},
}

\bib{Alpha}{article}{
      author={Jumper, John},
      author={Evans, Richard},
      author={Pritzel, Alexander},
      author={Green, Tim},
      author={Figurnov, Michael},
      author={Ronneberger, Olaf},
      author={Tunyasuvunakool, Kathryn},
      author={Bates, Russ},
      author={\v{Z}\'{i}dek, Augustin},
      author={Potapenko, Anna},
      author={Bridgland, Alex},
      author={Meyer, Clemens},
      author={Kohl, Simon A.~A.},
      author={Ballard, Andrew~J.},
      author={Cowie, Andrew},
      author={Romera-Paredes, Bernardino},
      author={Nikolov, Stanislav},
      author={Jain, Rishub},
      author={Adler, Jonas},
      author={Back, Trevor},
      author={Petersen, Stig},
      author={Reiman, David},
      author={Clancy, Ellen},
      author={Zielinski, Michal},
      author={Steinegger, Martin},
      author={Pacholska, Michalina},
      author={Berghammer, Tamas},
      author={Bodenstein, Sebastian},
      author={Silver, David},
      author={Vinyals, Oriol},
      author={Senior, Andrew~W.},
      author={Kavukcuoglu, Koray},
      author={Kohli, Pushmeet},
      author={Hassabis, Demis},
       title={Highly accurate protein structure prediction with {AlphaFold}},
        date={2021},
     journal={Nature},
      volume={596},
       pages={583\ndash 589},
         url={https://doi.org/10.1038/s41586-021-03819-2},
}

\bib{Risi2}{inproceedings}{
      author={Kondor, Risi},
      author={Lin, Zhen},
      author={Trivedi, Shubhendu},
       title={Clebsch--{G}ordan nets: a fully {Fourier} space spherical
  convolutional neural network},
        date={2018},
   booktitle={{Advances in Neural Information Processing Systems (NeurIPS)}},
      editor={Bengio, S.},
      editor={Wallach, H.},
      editor={Larochelle, H.},
      editor={Grauman, K.},
      editor={Cesa-Bianchi, N.},
      editor={Garnett, R.},
      volume={31},
       pages={10117\ndash 10126},
}

\bib{AlexNet}{inproceedings}{
      author={Krizhevsky, Alex},
      author={Sutskever, Ilya},
      author={Hinton, Geoffrey~E.},
       title={{ImageNet} classification with deep convolutional neural
  networks},
        date={2012},
   booktitle={{Advances in Neural Information Processing Systems (NIPS)}},
      editor={Pereira, F.},
      editor={others},
       pages={1097\ndash 1105},
}

\bib{Risi1}{inproceedings}{
      author={Kondor, Risi},
      author={Trivedi, Shubhendu},
       title={On the generalization of equivariance and convolution in neural
  networks to the action of compact groups},
        date={2018},
   booktitle={{Proceedings of the 35th International Conference on Machine
  Learning (ICML)}},
      editor={Dy, Jennifer},
      editor={Krause, Andreas},
      series={Proceedings of Machine Learning Research},
      volume={80},
       pages={2747\ndash 2755},
}

\bib{L}{article}{
      author={Lim, Lek-Heng},
       title={Tensors in computations},
        date={2021},
     journal={Acta Numer.},
      volume={30},
       pages={555\ndash 764},
         url={https://doi.org/10.1017/S0962492921000076},
}

\bib{Maron}{inproceedings}{
      author={Maron, Haggai},
      author={Fetaya, Ethan},
      author={Segol, Nimrod},
      author={Lipman, Yaron},
       title={On the universality of invariant networks},
        date={2019},
   booktitle={{Proceedings of the 36th International Conference on Machine
  Learning (ICML)}},
      editor={Chaudhuri, Kamalika},
      editor={Salakhutdinov, Ruslan},
      series={Proceedings of Machine Learning Research},
      volume={97},
       pages={4363\ndash 4371},
}

\bib{MP}{article}{
      author={McCulloch, Warren~S.},
      author={Pitts, Walter},
       title={A logical calculus of the ideas immanent in nervous activity},
        date={1943},
     journal={Bull. Math. Biophys.},
      volume={5},
       pages={115\ndash 133},
}

\bib{Rosenblatt}{article}{
      author={Rosenblatt, Frank},
       title={The perceptron: A probabilistic model for information storage and
  organization in the brain},
        date={1958},
     journal={Psychol. Rev.},
      volume={65},
       pages={386\ndash 408},
}

\bib{townshend_RNA_2021}{article}{
      author={Townshend, Raphael J.~L.},
      author={Eismann, Stephan},
      author={Watkins, Andrew~M.},
      author={Rangan, Ramya},
      author={Karelina, Maria},
      author={Das, Rhiju},
      author={Dror, Ron~O.},
       title={Geometric deep learning of {RNA} structure},
        date={2021},
     journal={Science},
      volume={373},
      number={6558},
       pages={1047\ndash 1051},
}

\bib{Tu}{article}{
      author={Tu, Loring~W.},
       title={What is {$\dots$} equivariant cohomology?},
        date={2011},
     journal={Notices Amer. Math. Soc.},
      volume={58},
      number={3},
       pages={423\ndash 426},
}

\bib{weiler_3d_2018}{inproceedings}{
      author={Weiler, Maurice},
      author={Geiger, Mario},
      author={Welling, Max},
      author={Boomsma, Wouter},
      author={Cohen, Taco~S.},
       title={{3D Steerable CNNs}: Learning rotationally equivariant features
  in volumetric data},
        date={2018},
   booktitle={{Advances in Neural Information Processing Systems (NeurIPS)}},
      volume={31},
   publisher={Curran Associates, Inc.},
       pages={10381\ndash 10392},
}

\bib{Zhang}{inproceedings}{
      author={Zhang, Richard},
       title={Making convolutional networks shift-invariant again},
        date={2019},
   booktitle={{Proceedings of the 36th International Conference on Machine
  Learning (ICML)}},
      editor={Chaudhuri, Kamalika},
      editor={Salakhutdinov, Ruslan},
      series={Proceedings of Machine Learning Research},
      volume={97},
       pages={7324\ndash 7334},
}

\bib{zaheer_deep_2017}{inproceedings}{
      author={Zaheer, Manzil},
      author={Kottur, Satwik},
      author={Ravanbakhsh, Siamak},
      author={Poczos, Barnabas},
      author={Salakhutdinov, Russ~R.},
      author={Smola, Alexander~J.},
       title={Deep sets},
        date={2017},
   booktitle={{Advances in Neural Information Processing Systems (NIPS)}},
      editor={Guyon, I.},
      editor={Luxburg, U.~Von},
      editor={Bengio, S.},
      editor={Wallach, H.},
      editor={Fergus, R.},
      editor={Vishwanathan, S.},
      editor={Garnett, R.},
      volume={30},
       pages={3391\ndash 3401},
}

\end{biblist}
\end{bibdiv}

\end{document}